\documentclass[journal]{IEEEtran}

\usepackage{verbatim}
\usepackage{amsfonts}
\usepackage{amssymb}
\usepackage{stfloats}
\usepackage{cite}
\usepackage{graphicx}
\usepackage{psfrag}
\usepackage{amsmath}
\usepackage{array}
\usepackage{epstopdf}
\usepackage{authblk}
\usepackage{graphicx} 
\usepackage{amsthm} 
\usepackage{lipsum}
\usepackage{verbatim} 
\usepackage{authblk}
\usepackage{mathtools}
\usepackage{cuted}
\usepackage{booktabs}

\usepackage{amsmath}
\usepackage{mathrsfs}

\usepackage{algorithmic}

\usepackage{array}

\ifCLASSOPTIONcompsoc
\usepackage[caption=false,font=normalsize,labelfont=sf,textfont=sf]{subfig}
\else
\usepackage[caption=false,font=footnotesize]{subfig}
\fi

\usepackage{url}
\usepackage{graphicx,amsmath,amssymb,amsfonts}
\usepackage{algorithmic,algorithm}

\usepackage{hyperref}
\hypersetup{colorlinks=true, citecolor=blue}

\usepackage{setspace}

\makeatletter

\newcommand{\Rmnum}[1]{\expandafter\@slowromancap\romannumeral #1@}
\makeatother

\usepackage{framed} 

\usepackage{cancel}

\usepackage{booktabs}
\usepackage{tabularx,makecell,multirow}

\usepackage[table]{xcolor}
\usepackage{pifont} 

\usepackage{diagbox}

\definecolor{shadecolor}{rgb}{0.92,0.92,0.92}

\begin{document}
	\bstctlcite{ref:BSTcontrol}
	
	\title{Federated Split Learning for Resource-Constrained Robots in Industrial IoT: Framework Comparison, Optimization Strategies, and Future Directions}
	
	\author{Wanli Ni, Hui Tian, Shuai Wang, Chengyang Li, Lei Sun, and Zhaohui Yang
		
	\vspace{-3 mm}		
	\thanks{Wanli Ni is with the Department of Electronic Engineering, Tsinghua University, Beijing 100084, China (e-mail: niwanli@tsinghua.edu.cn).}
	\thanks{Hui Tian is with the State Key Laboratory of Networking and Switching Technology, Beijing University of Posts and Telecommunications, Beijing 100876, China (email: tianhui@bupt.edu.cn).}
	\thanks{Shuai Wang is with the Shenzhen Institutes of Advanced Technology, Chinese Academy of Sciences, Shenzhen, China (email: s.wang@siat.ac.cn).}	
	\thanks{Chengyang Li is with the Department of Electrical and Electronic Engineering, The University of Hong Kong, Hong Kong, China. (chengyangli@connect.hku.hk).}	
	\thanks{Lei Sun is with the School of Automation and Electrical Engineering, University of Science and Technology Beijing, Beijing 100083, China (email: sun\_lei@ustb.edu.cn).}
	\thanks{Zhaohui Yang is with the Zhejiang Lab, Hangzhou 311121, China, and also with the College of Information Science and Electronic Engineering, Zhejiang University, Hangzhou, Zhejiang 310027, China (yang\_zhaohui@zju.edu.cn).}
}
	
	\maketitle
	
	\begin{abstract}
	Federated split learning (FedSL) has emerged as a promising paradigm for enabling collaborative intelligence in industrial Internet of Things (IoT) systems, particularly in smart factories where data privacy, communication efficiency, and device heterogeneity are critical concerns.
	In this article, we present a comprehensive study of FedSL frameworks tailored for resource-constrained robots in industrial scenarios.
	We compare synchronous, asynchronous, hierarchical, and heterogeneous FedSL frameworks in terms of workflow, scalability, adaptability, and limitations under dynamic industrial conditions. Furthermore, we systematically categorize token fusion strategies into three paradigms: input-level (pre-fusion), intermediate-level (intra-fusion), and output-level (post-fusion), and summarize their respective strengths in industrial applications.
	We also provide adaptive optimization techniques to enhance the efficiency and feasibility of FedSL implementation, including model compression, split layer selection, computing frequency allocation, and wireless resource management. Simulation results validate the performance of these frameworks under industrial detection scenarios.
	Finally, we outline open issues and research directions of FedSL in future smart manufacturing systems.
	\end{abstract}
	

	\vspace{-2 mm}
	\section{Introduction}
	The rapid evolution of the industrial Internet of Things (IoT) has catalyzed a paradigm shift toward intelligent, autonomous, and interconnected manufacturing systems \cite{Sisinni2018Industrial}.
	At the heart of this transformation are networked robots that perform complex tasks such as quality inspection, predictive maintenance, and multi-device collaboration across dynamic production environments.
	These robots are increasingly equipped with multimodal sensors and onboard computing units, enabling them to perceive, reason, and act in real time \cite{Hu2025Wireless}.
	However, the deployment of data-driven artificial intelligence (AI) models in such settings faces critical challenges, including data privacy concerns, limited communication bandwidth, heterogeneous hardware capabilities, and stringent latency requirements.
	
\begin{figure*}[t]
		\centering
		\includegraphics[width=7 in]{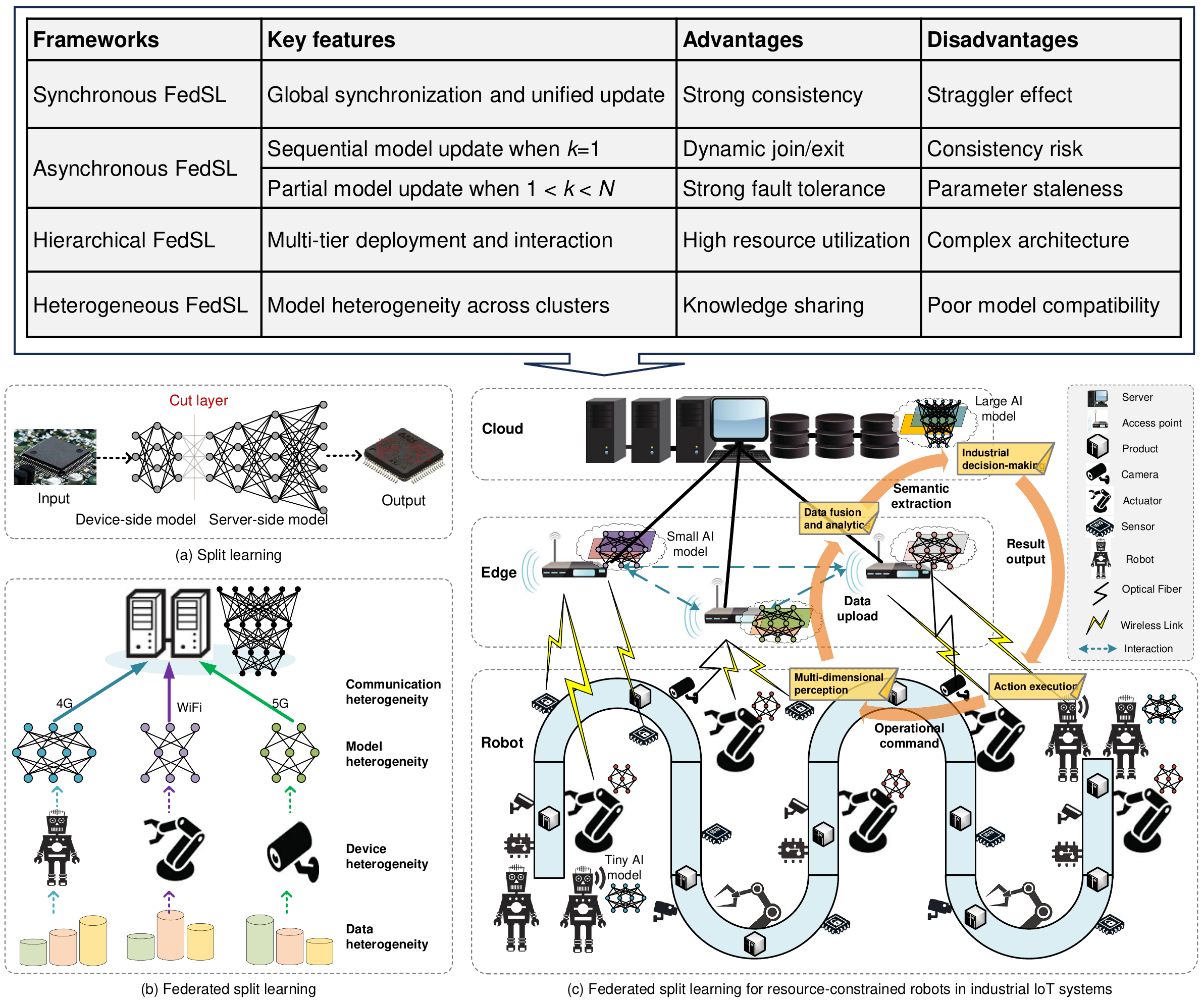}
		\caption{(a) An illustration of split lerning. (b) The implementation of vanilla  FedSL in industrial environments faces several key technical challenges, including data heterogeneity, device heterogeneity, model heterogeneity, and communication heterogeneity. i) Data heterogeneity arises from the imbalance data across devices, where local datasets may exhibit significant statistical divergence due to varying usage patterns or environments. ii) Device heterogeneity refers to the disparities in computational power, memory capacity, and energy constraints among robots. iii) Model heterogeneity occurs when personalized AI models with different architectures or parameter dimensions are deployed across devices. iv) Lastly, communication heterogeneity stems from the variability in network connectivity, such as 4G, 5G, and Wi-Fi, resulting in fluctuating bandwidth, latency, and reliability during model updates and intermediate data transmission.
		(c) FedSL for resource-constrained robots in industrial IoT systems within a three-tier robot-edge-cloud architecture. 
		In this article, we use the terms ``device", ``client" and ``robot" interchangeably, which does not affect the readability.}
		\label{Fig1}
	\end{figure*}

	Traditional centralized learning approaches, which involve collecting raw data on a central server for model training, are not well-suited for industrial IoT systems due to increasing data volumes, high transmission latency, and strict regulations on data sharing.
	Federated learning has emerged as a privacy-preserving alternative, allowing distributed devices to collaboratively train a global model without exchanging local data \cite{mcmahan2017fl}.
	However, in resource-constrained robotic systems, federated learning still incurs significant communication overhead due to the transmission of full model gradients, and places heavy computational burdens on edge devices during local training.
	To address these limitations, federated split learning (FedSL) has recently gained attention as a hybrid learning paradigm that combines the strengths of split learning and federated learning~\cite{Thapa2022SplitFed}.
	As shown in Fig. \ref{Fig1}, deep neural networks are partitioned between client devices (e.g., robots) and edge servers in FedSL, with devices computing partial forward propagation up to a predefined split layer and transmitting intermediate features to the server for completion of one iteration~\cite{Ni2024FedSL}.
	This approach significantly reduces the computational load on client devices and enhances privacy by preventing raw data from leaving the robot. Moreover, by decoupling model execution across heterogeneous participants, FedSL enables fine-grained adaptation to varying communication, computation, and energy constraints across industrial robots~\cite{Lin2025AdaptSFL}.
	Although FedSL shows potential, the application of FedSL in industrial robotics remains fragmented and insufficiently studied. Existing studies often focus on isolated aspects rather than providing a comprehensive analysis of architectural diversity, multimodal fusion strategies, or cross-layer optimization techniques tailored to industrial tasks \cite{Xu2024Accelerating, Ao2025Semi, Liang2025Communication}.
	Furthermore, the heterogeneity of robotic platforms (in terms of sensors, computing power, and task objectives) demands flexible FedSL frameworks that go beyond one-size-fits-all designs.
	
	In this article, we present a comprehensive analysis of FedSL frameworks in the context of networked robots towards industrial intelligence.
	The primary contributions of this paper are threefold:
	We begin by reviewing and comparing four representative FedSL frameworks (synchronous, asynchronous, hierarchical, and heterogeneous FedSL), highlighting their operational workflows, scalability, robustness to network dynamics, and suitability for different industrial tasks.
	Subsequently, we provide a taxonomy of token fusion strategies for multimodal data in industrial environments, such as pre-fusion (early fusion), intra-fusion (intermediate fusion), and post-fusion (late fusion) paradigms, and then analyze their performance trade-offs in real-world tasks such as vision-based inspection, human–robot interaction, and equipment maintenance.
	In addition, we discuss adaptive optimization techniques that can enhance the efficiency of FedSL frameworks, including split layer selection, resource management, and AI model compression. Simulation results validate the performance of these frameworks under resource-limited industrial conditions.
	Finally, we list several future research directions for deploying FedSL in smart factories, such as dynamic task adaptation, human-in-the-loop collaboration, and energy-aware scheduling.
	
	The remainder of this paper is organized as follows: 
	Section~II presents a comparative analysis of FedSL frameworks and introduces token fusion strategies for networked robots. 
	Section III discusses adaptive optimization strategies used in FedSL.
	Section IV provides simulation results for FedSL in industrial tasks.
	Section V outlines challenges and future directions, which is followed by the conclusion in Section VI.

	\section{FedSL Frameworks for Industrial IoT}
	As illustrated in Fig.~\ref{Fig1}, unlike federated learning, which trains complete models on edge devices, FedSL combines model splitting with federated aggregation, enabling resource-constrained IoT nodes to collaboratively train complex AI models without sharing raw data.
	However, the integration of FedSL into industrial systems requires tailored frameworks that balance computational efficiency, communication overhead, and model accuracy, particularly for industries reliant on heterogeneous robots.	
	This section provides an in-depth analysis of four FedSL frameworks (i.e., asynchronous, synchronous, hierarchical, and heterogeneous FedSL), each tailored to address distinct challenges in industrial environments.

	\begin{figure*}[t]
		\centering
		\includegraphics[width=7.0 in]{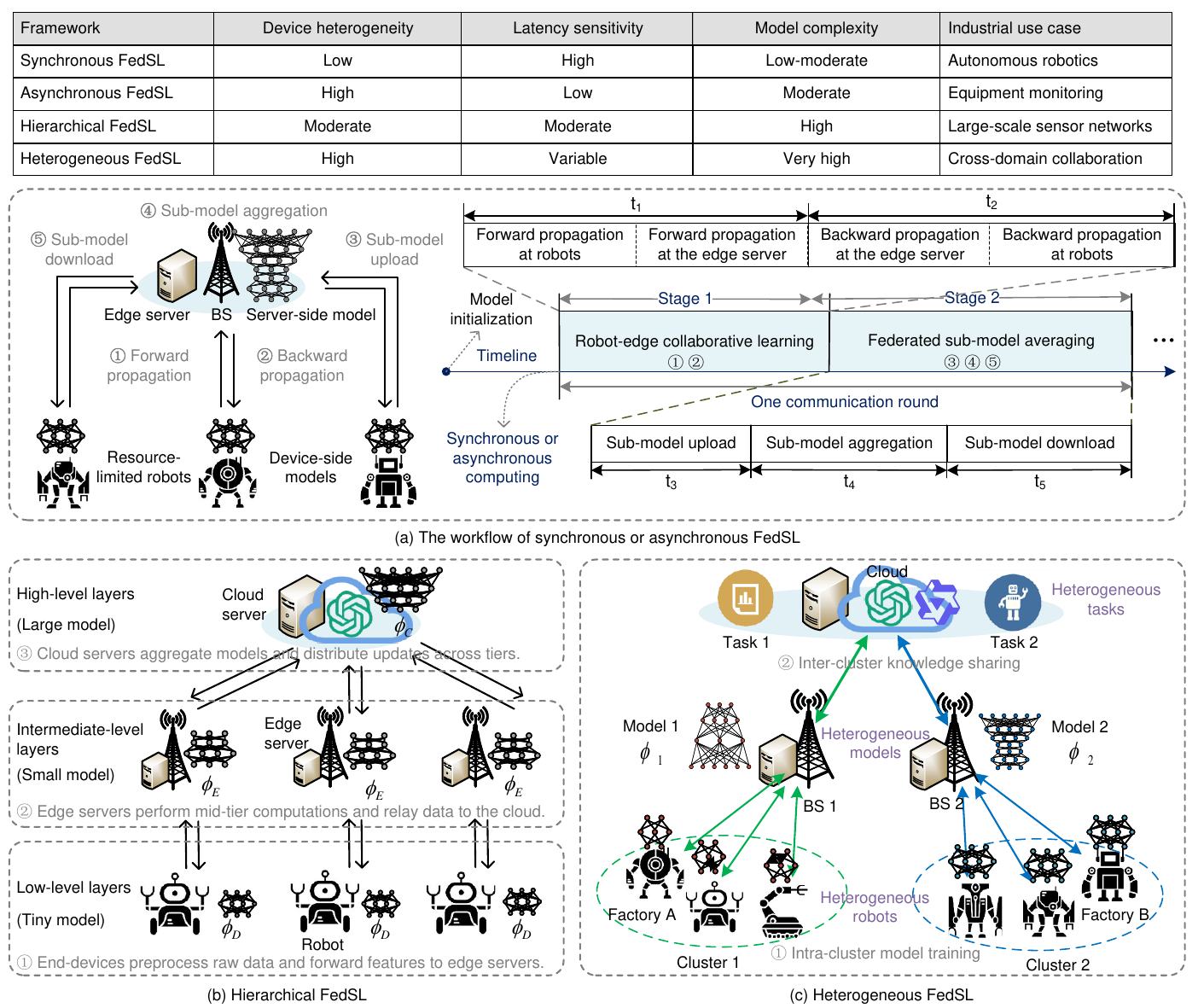}
		\caption{Framework comparison of different FedSL paradigms. (a) An illustration of the workflow of synchronous or asynchronous FedSL. (b) Hierarchical FedSL, which decomposes large-scale AI models into three tiers. End devices are responsible for lightweight feature extraction, edge servers handle intermediate computational tasks, and the cloud performs high-level, complex reasoning processes. (c) Heterogeneous FedSL, which addresses model heterogeneity across IIoT clusters (e.g., different factories or product lines). Within each cluster, a specialized model is trained using the standard FedSL approach, while cross-cluster knowledge integration is achieved through cloud-based distillation and sharing mechanisms.}
		\label{Fig2}
	\end{figure*}
	
	\subsection{Comparison of Typical FedSL Frameworks}	
	In this subsection, we compare four representative FedSL frameworks in terms of their fundamental mechanisms, strengths, and limitations.
	In the following, we discuss the workflow designs of different FedSL frameworks and their potentials for industrial implementation.
	
	\textbf{Synchronous FedSL Framework:}
	Synchronous FedSL enforces global synchronization, requiring all clients to complete forward/backward propagation before the server aggregates device-side models \cite{Ni2024FedSL}.
	This framework operates in lockstep cycles. All clients initiate training simultaneously, using private datasets to update the device-side model parameters. After completing local iterations, clients submit their model parameters to the server, which waits until all contributions are received. The server then computes a weighted average and updates the global model. This cycle continues until predefined convergence criteria are met.
	Synchronous FedSL is ideal for synchronized manufacturing lines, such as automotive assembly plants where robotic arms must coordinate movements with millisecond precision. For example, in a paint shop, cameras on robotic arms capture images of car bodies, and a model synchronously trained across cameras could ensure consistent color matching by aggregating gradients from all devices simultaneously.
	However, their reliance on full client participation makes them vulnerable to stragglers, e.g., slow or offline devices that delay the entire workflow.

	\textbf{Asynchronous FedSL Framework:}
	Unlike synchronous approaches, asynchronous FedSL allows clients to participate in model updates at their own pace.
	Two operational modes dominate this category: sequential client interaction ($k=1$) and threshold-based aggregation ($1 < k < N$), where $k$ represents the number of clients participating in the model aggregation per period, and $N$ is the total number of clients.
	\begin{itemize}
		\item 
		For the sequential client interaction ($k=1$), clients interact with an edge server in a strict sequence.
		The server updates the model using the raw data from a single client before broadcasting the updated parameters to the next client in the queue.
		However, frequent model updates from individual clients can destabilize convergence, as the server-side model’s direction may shift abruptly with each client’s contribution.
		Additionally, if a client fails mid-training (e.g., due to network disconnection or hardware malfunction), it can block subsequent clients.
		\item 
		For the threshold-based aggregation ($1 < k < N$), it adopts a dynamic aggregation criterion. Here, the server aggregates gradients from the first $k$ clients that meet predefined thresholds, such as completing a minimum number of local iterations or operating within a specified time window \cite{Ao2025Semi}.
		Thus, the choice of $k$ and threshold values significantly impacts model performance.
	\end{itemize}
	Asynchronous FedSL often excels in industrial environments with non-uniform device capabilities. For instance, in a factory with legacy machines and modern IoT robots, slower devices can contribute gradients at their own pace without delaying faster nodes.
	Despite its advantages, asynchronous FedSL faces challenges like gradient staleness, where outdated parameters degrade model convergence and accuracy.

	\textbf{Hierarchical FedSL Framework:}
	Hierarchical FedSL partitions a deep neural network into multiple parts, distributing computational loads across local robots, edge servers, and the cloud, as illustrated in Fig. \ref{Fig2}a.
	This framework leverages the proximity of edge servers to industrial assets while offloading complex computations to more powerful infrastructure \cite{Guo2024Hierarchical}.
	The workflow begins with end-devices preprocessing raw data and extracting low-dimensional features. These features are transmitted to edge servers, which perform mid-tier computations. Edge servers then forward processed data to the cloud, where high-level complex reasoning occurs. The cloud distributes updated parameters back to edge and end-devices, completing the feedback loop.
	Hierarchical FedSL is particularly effective in large-scale industrial deployments, such as oil refineries or logistics hubs, where thousands of sensors generate high-velocity data. 
	However, hierarchical frameworks requires domain expertise to balance communication overhead (e.g., feature transmission between layers) and computational load. Moreover, single points of failure in edge or cloud layers can disrupt entire workflows, necessitating redundancy and failover mechanisms.
	
	\textbf{Heterogeneous FedSL Framework:}
	Heterogeneous FedSL addresses the diversity of AI models and tasks across industrial subsystems, as illustrated in Fig. \ref{Fig2}b.
	Unlike homogeneous FedSL, which assumes uniform model architectures, heterogeneous approaches support cross-cluster collaboration where different clusters (e.g., factories, product lines) train specialized models, and knowledge is shared across clusters through the cloud \cite{Liao0XYHQ24}.
	Specifically, intra-cluster training begins with cluster-specific model initialization. Robots within the cluster perform training on private data, transmitting smashed data to a cluster-level server for collaboration. Upon convergence, the cloud facilitates inter-cluster sharing through mechanisms like knowledge distillation or parameter sharing. Alternatively, non-sensitive parameters (e.g., batch normalization layers) may be shared directly. 
	Heterogeneous FedSL is well-suited for multi-task ecosystems, such as automotive manufacturers coordinating across engine assembly, painting, and quality control clusters. For example, a cluster training a model for engine defect detection could share anomaly patterns with a maintenance cluster, enabling predictive repairs.
	However, in heterogeneous FedSL, the effectiveness of knowledge distillation depends on the similarity between teacher and student tasks; for instance, distilling knowledge from a vision model to a time-series model may yield limited gains.
	
	\begin{shaded} 
	In summary, FedSL offers a versatile toolkit for deploying AI models in resource-constrained industrial systems, addressing challenges such as device heterogeneity, data privacy, and edge-end collaboration.
	In Fig. \ref{Fig2}, we compare various FedSL frameworks in terms of latency, model complexity, and use case.
	Specifically, synchronous FedSL ensures reliability in time-critical scenarios, asynchronous FedSL excels in dynamic environments, hierarchical FedSL optimizes resource usage, and heterogeneous FedSL enables cross-domain knowledge sharing. 
	Selecting the optimal FedSL framework for realistic production environments requires balancing device capabilities, latency requirements, and task complexity carefully.
	\end{shaded} 

	\subsection{Token Fusion for Networked Robots with Multimodal Data}
	It can be foreseeable that the extensive application of large models in industrial robots brings unprecedented opportunities, but this also presents significant challenges in handling multimodal data from heterogeneous sensors such as cameras, force/torque sensors, and environmental monitors.
	Token fusion determines when and how multimodal sensor data is integrated into a unified representation for downstream tasks.
	To make full use of these multimodal data, this subsection introduces three token fusion approaches for intelligent robots to balance the trade-offs between accuracy, latency, and computational efficiency, enabling large models to operate effectively in resource-constrained industrial systems.

\begin{figure*}[t]
		\centering
		\includegraphics[width=7.0 in]{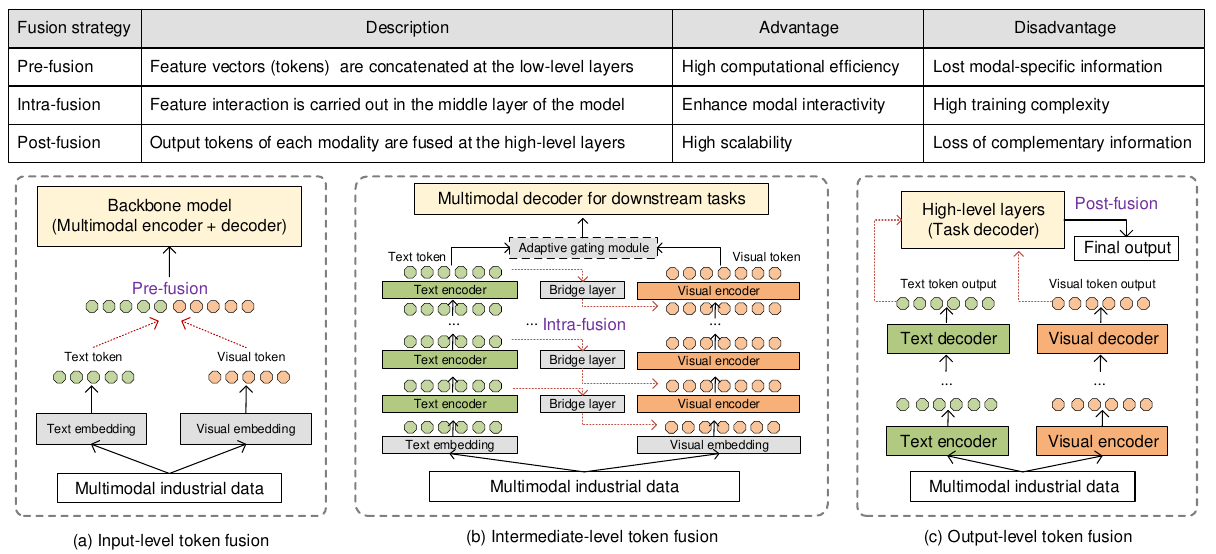}
		\caption{Three token fusion strategies: (a) input-level token fusion, (b) intermediate-level token fusion, and (c) output-level token fusion.}
		\label{Fig3_token_fusion}
	\end{figure*}

	\textbf{Pre-Fusion (Early Fusion):}
	As illustrated in Fig. \ref{Fig3_token_fusion}a, pre-fusion involves concatenating tokens from multiple modalities (e.g., text instructions, camera pixels, LiDAR point clouds) at the input layer of the backbone model. This enables the model to learn joint representations from the early stages, making it particularly effective for tasks requiring tight cross-modal coordination, such as robotic manipulation or autonomous navigation.
	To ensure compatibility, modality-specific embedding layers are typically applied to normalize token dimensions before concatenation \cite{Zhang2025Token}.
	When using pre-fusion in FedSL, the device-side model need to process and combine raw multimodal data (e.g., aligning camera frames with robotic arm positions) into a unified token sequence prior to sending it to the backbone model that runs on the server side.

	\textbf{Intra-Fusion (Intermediate Fusion):}
	Intra-fusion integrates tokens at intermediate layers of the backbone model (e.g., after early self-attention blocks in a transformer), enabling contextual interaction while preserving modality-specific representations \cite{Shi2025SwimVG}, as shown in Fig.~\ref{Fig3_token_fusion}b.
	This fusion is typically achieved through cross-attention mechanisms or adaptive gating modules that dynamically weigh the contribution of each modality based on context.
	This approach suits tasks requiring fine-grained spatiotemporal understanding, such as detecting subtle defects on fast-moving conveyor belts.
	When applying intra-fusion in hierarchical FedSL, devices extract initial features, then transmit intermediate representations to an edge server. Token fusion occurs at this intermediate tier, where shared context is constructed before further processing in the cloud.

	\textbf{Post-Fusion (Late Fusion):}
	When applying post-fusion in heterogeneous FedSL, task-oriented or modality-specific models are trained within individual clusters. The predicted token outputs or high-level embeddings from these models are aggregated in the cloud using ensemble techniques, such as cross-model knowledge distillation.
	This approach is particularly suitable for applications tolerating delayed decisions, such as predictive maintenance or quality root-cause analysis.
	From Fig. \ref{Fig3_token_fusion}c, by integrating the outputs of different models, it can be inferred that post-fusion enables knowledge transfer across modalities, domains and factories. For example, a vision-based model (trained on factory A's image data) can share learned fault patterns with a text-based model (trained on factory B's log data), even if the two models differ in architecture and input modality.

	\begin{shaded} 
		According to the specific task requirements and hardware conditions, industrial practitioners should employ appropriate token fusion strategies to process multimodal sensor data efficiently.
		Specifically, pre-fusion enhances early feature integration, making it well-suited for tasks requiring holistic scene understanding.
		Intra-fusion enables real-time adaptability through partial cross-modal interaction and is suitable for complex industrial scenarios.
		Post-fusion mitigates modality-specific noise by delaying integration until decision-level outputs, which benefits fault-tolerant applications and facilitates the incorporation of new data modalities.
	\end{shaded} 
	\vspace{-3 mm}
	
	\section{Adaptive Optimization Techniques for FedSL}
	This section explores adaptive optimization techniques to tackle challenges related to computational load balancing, communication efficiency, and model adaptability, aiming to enable efficient and scalable FedSL implementations.

	\subsection{Optimization of AI Model Architecture}
	The architecture of AI models significantly impacts the overall FedSL performance. Techniques such as quantization and pruning are commonly employed to compress models, thereby reducing the data transmitted between clients and servers while accelerating local computation \cite{Zhang2025Federated}.
	However, aggressive compression may introduce quantization errors or feature loss, which may degrade model accuracy, especially when clients have limited data for fine-tuning. For instance, when pruning vision transformers deployed in edge-based defect detection systems, it is essential to strike a balance between achieving sparsity and maintaining robustness against sensor noise. Consequently, optimizing model architectures in FedSL necessitates a careful trade-off among compression-induced accuracy degradation, hardware-specific limitations, and the dynamic nature of data distributions.
	In addition, static pruning masks derived from centralized training become suboptimal as client data distributions shift.
	Combining quantization-aware training with adaptive precision scaling (e.g., mixed-precision quantization) and developing dynamic pruning methods robust to imbalance data distributions are critical for FedSL implemented in resource-limited factories.

	\subsection{Optimization of Split Layer Selection}
	The split layer determines the division of a deep neural network between local robots (processing shallow layers) and edge servers (handling deep layers). A suboptimal split layer can lead to excessive client-side computation (overloading low-power devices) or high communication overhead (transmitting large intermediate features). For example, in drone-based surveillance systems, splitting too early may require transmitting high-dimensional feature maps, overwhelming wireless bandwidth. Conversely, splitting too late may strain client resources, making it unsuitable for battery-powered drones. Thus, selecting the optimal split layer is pivotal to balance latency, energy consumption, and privacy preservation \cite{Lin2025Hierarchical}.
	However, the adaptive optimization of split layer in FedSL encounters several challenges, notably the heterogeneity of device capabilities and the trade-off between privacy and utility. Specifically, devices differ in terms of computational resources, such as central processing units (CPUs) versus graphics processing units (GPUs), memory capacity, and battery life. Furthermore, shallow splits may leak sensitive intermediate features, while deep splits enhance privacy but increase device computational load.

	\subsection{Optimization of Computing Frequency}  
	Clients in FedSL are required to execute local computations, such as forward and backward propagation. The dynamic adjustment of computing frequency, including CPU and GPU clock speeds, serves to minimize energy consumption while ensuring that processing deadlines are met. For instance, in wearable health monitoring systems, excessive frequency scaling can result in delayed critical notifications, such as fall detection alerts, whereas insufficient frequency scaling may lead to rapid battery depletion.
	Moreover, although higher frequencies reduce computation time, they increase power consumption disproportionately. Prolonged operation at elevated frequencies can also cause device overheating, which may activate thermal throttling mechanisms.
	Consequently, integrating dynamic voltage and frequency scaling with predictive task scheduling, alongside the exploration of hardware accelerators, represents a promising approach for achieving energy-efficient computations in FedSL frameworks.
	
	\subsection{Optimization of Radio Resource Allocation}
	FedSL requires frequent transmission of smashed data (intermediate features) between clients and servers. In wireless networks, channel conditions vary due to fading, interference, and mobility, leading to packet loss and retransmissions. Efficient radio resource allocation (e.g., time, spectrum, and power) helps to minimize latency and energy consumption while ensuring reliable communication \cite{Liang2025Communication}. For instance, in smart factories, robots must transmit sensor data to edge servers with sub-millisecond latency for real-time control, necessitating optimal resource allocation.
	Nevertheless, the presence of obstacles and interference in time-varying wireless links poses significant challenges to the optimization of radio resource allocation, particularly in scenarios involving battery-powered devices with constrained energy capacities. Furthermore, accommodating the simultaneous connectivity of hundreds of clients necessitates the implementation of adaptive scheduling strategies and robust interference management techniques.

	\begin{shaded} 
		A key takeaway is that adaptability is crucial in FedSL, given that static configurations tend to underperform when confronted with heterogeneous device capabilities, fluctuating network conditions, and diverse task requirements. 
		Although dynamic methods may bring significant gains but often at the cost of increased complexity. There is no one-size-fits-all solution in FedSL; instead, optimal performance emerges from carefully balancing competing objectives, thus meeting the multifaceted demands of real-world deployments.
	\end{shaded} 
	\vspace{-3 mm}

	\begin{figure}[t]
		\centering
		{\subfloat[Training loss vs. communication rounds.]{
				\includegraphics[width=3.1 in]{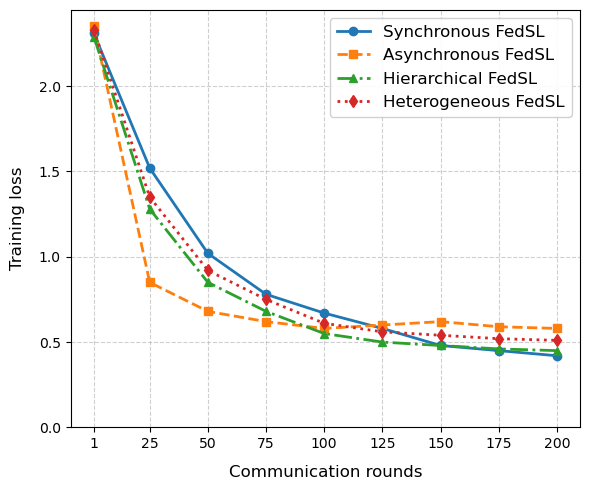} \label{Fig4a}
			}
		}
		{\subfloat[Test accuracy vs. packet loss rate.]{
				\includegraphics[width=3.1 in]{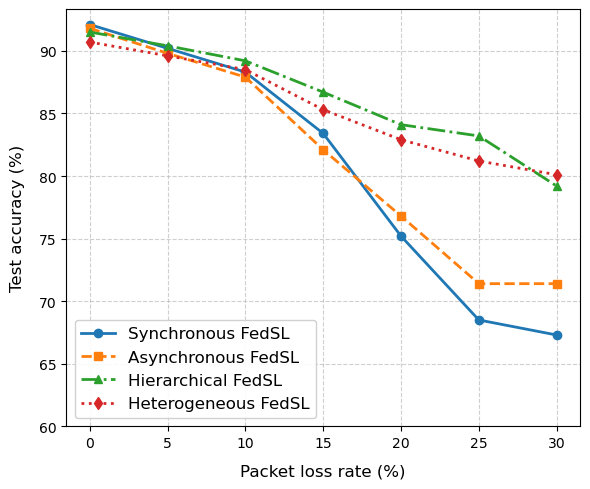} \label{Fig4b}
			}
		}
		\caption{Training performance of FedSL frameworks in industrial systems.} 
		\label{Fig4}
		\vspace{-3 mm}
	\end{figure}

	\section{Simulation Results}
	In this section, we evaluate the performance of various FedSL frameworks in an industrial IoT scenario involving networked robots.	
	The simulation is conducted in a $50 \text{m} \times 50 \text{m}$ warehouse environment having $N=10$ robots and an edge server.
	Communications between robots and the edge server occurs over a wireless channel with the following parameters: a bandwidth of $10$ MHz, a transmission power of $23$ dBm, and a noise power spectral density of $-85$ dBm.
	The communication model accounts for path loss and random packet loss to simulate unstable industrial wireless conditions.
	Each robot is equipped with a VGG16 model for real-time parcel detection.
	The VGG16 is partitioned at layer 7: local robots perform forward and backward propagation for layers 1–7, while the edge server handles layers 8–16 and computes the final loss.
	We assume the parcel dataset is uniformly distributed across all robots.
	Model training employs the cross-entropy loss function, with 5 local iterations per communication round and a fixed learning rate of 0.001.
	Furthermore, four FedSL frameworks are set as follows:
	i) Synchronous FedSL: Global model aggregation is performed every 25 communication rounds to synchronize their updates.
	ii) Asynchronous FedSL: Model updates are aggregated as soon as any group of $k=5$ robots completes the preset number of local iterations.
	iii) Hierarchical FedSL: The VGG16 model is partitioned across three tiers. Layers 1–3 are executed on the robot, layers 4–7 on the edge server, and layers 8–16 on the cloud.
	iv) Heterogeneous FedSL: All robots are divided into two clusters of 5 robots each. Cluster 1 trains a VGG16 for parcel detection, while Cluster 2 trains a ResNet-18 using industrial defect images, in which the ResNet-18 is partitioned at layer 6.

	In Fig. \ref{Fig4}, we demonstrate the training performance of four FedSL frameworks across 200 communication rounds.
	Fig. \ref{Fig4a} illustrates that synchronous FedSL exhibits a steady but relatively slow convergence rate.
	Asynchronous FedSL demonstrates a rapid initial descent in training loss, showcasing its ability to quickly adapt and learn from the data. However, this framework suffers from volatility due to stale parameter updates, as exemplified at communication round 150, where there is a noticeable increase in loss. 
	Hierarchical FedSL achieves superior convergence stability by reducing global communication variance through its multi-tier model deployment and interaction strategy. It outperforms synchronous FedSL before communication round 120 and continues to maintain a stable and low loss value until round 200.
	Heterogeneous FedSL maintains a stable descent, and achieves a stable loss value at communication round 175, though slightly higher than hierarchical FedSL.
	In summary, each FedSL framework has unique strengths and weaknesses. Synchronous FedSL offers steady convergence but is sensitive to stragglers. Asynchronous FedSL provides rapid initial learning but suffers from parameter staleness. Hierarchical FedSL excels in stability and efficiency, making it suitable for large-scale distributed systems. Heterogeneous FedSL demonstrates robustness in handling diverse models, albeit with a slightly higher final loss.
	Fig. \ref{Fig4b} demonstrates the test accuracy under varying packet loss rates for four different FedSL frameworks.
	Synchronous FedSL exhibits a sharp degradation in performance as the packet loss rate increases.
	This decline can be attributed to the strict synchronization requirements of this framework, which are highly sensitive to network disruptions and packet losses.
	Asynchronous FedSL shows greater resilience to packet loss compared to the synchronous approach.
	It sustains a test accuracy of 71\% at a 30\% packet loss rate by tolerating stale updates.
	Hierarchical FedSL mitigates the impact of packet loss through end-edge-cloud collaboration. By leveraging a multi-tier architecture, it maintains a test accuracy of 79\% at a 30\% packet loss rate.
	Heterogeneous FedSL achieves a test accuracy of 80\% at a 30\% packet loss rate. By adapting to the heterogeneity of models and network environments, this framework can better handle packet loss and maintain high performance.

\begin{figure*}[t]
	\centering
	\includegraphics[width=7 in]{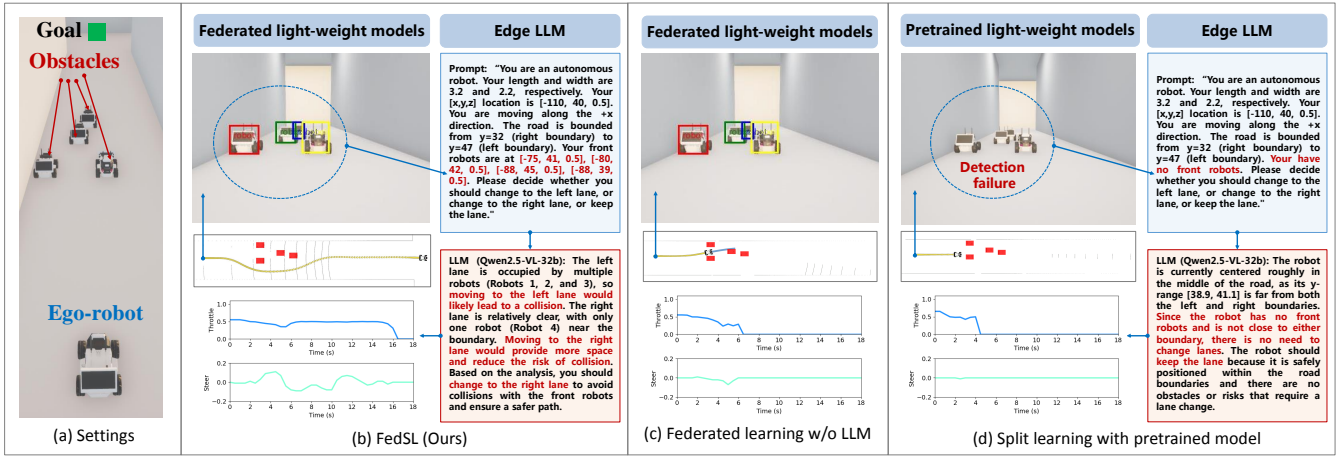}
	\caption{Perception, decision, and control profiles of different schemes in a Carla-based indoor factory simulation. (a) Scenario setup with an ego-robot navigating a corridor with 4 obstacles. (b) FedSL enables accurate obstacle detection and safe right detour via federated models and LLM-based decision-making. (c) Pure federated learning without high-level LLM reasoning. (d) Pure split learning with pretrained model.}
	\label{Fig5}
\end{figure*}

	In Fig. \ref{Fig5}, we evaluate the proposed FedSL in Carla simulation using Python and ROS.
	As shown in Fig. \ref{Fig5}a, we consider an indoor factory scenario with 4 abnormal obstacles. The ego-robot needs to pass the corridor, by reading sensor data (i.e., camera images and lidar points) and generating control actions (i.e., throttle and steer). To accomplish this task, the computation model consists of perception, decision, and control blocks.
	As shown in Fig. \ref{Fig5}b, we split the computation model into two parts, where the light-weight part (i.e., perception and control blocks) is deployed at the robot and the heavy part (i.e., the decision block) is deployed at the edge. Particularly, we finetune Yolo in this scenario using two-robot federated learning and adopt the federated Yolo for perception. The generated topology abstracts are fed to a large language model (LLM), i.e., Qwen2.5-VL-32b, for making decisions (i.e., left lane-change, right lane-change, or lane-keeping), which are then passed to the control block for taking actions. It can be seen that our FedSL detects all the abnormal obstacles accurately, takes a detour to the right lane, and successfully passes the corridor. Furthermore, throughout the entire computation, no sensor data is uploaded to the edge server, thereby protecting the privacy of the factory. 
	As shown in Fig. \ref{Fig5}c, the pure federated learning scheme, though detecting the obstacles, turns left for a shorter path, which fails in reasoning about the future risk without LLM guidance. Thus, the robot gets stuck inside the crowd.
	The pure split learning scheme adopts a pretrained Yolo, which fails in detecting the obstacles. Hence, the abstraction uploaded to the edge is incorrect, misleading the LLM towards a wrong decision. Thus, the robot keeps its current lane and gets stuck by a front robot, as shown in Fig.~\ref{Fig5}d.
	The above results demonstrate that FedSL enhances the robot capabilities by leveraging both the knowledge of other robots and the computation power of the edge, without any privacy leakage.

	\section{Challenges and Future Directions}
	
	\subsection{Adaptive FSL for Dynamic Industrial Environments}
	Current FSL implementations have difficulty handling the dynamic nature of modern manufacturing systems. Industrial robots often face: i) changes in tasks, such as alternating between welding and assembly, ii) modifications in hardware, like replacing robotic arms, and iii) environmental fluctuations, for example, temperature changes that impact sensor data. Conventional static model partitioning methods are ineffective in these scenarios, resulting in reduced performance or communication delays.
	Future research directions should focus on: 1) Developing meta-learning-based split architecture capable of rapid adaptation through few-shot updates. 2) Designing neural architecture search methods tailored for devices with limited resources, allowing automatic modification of AI models in response to real-time computational demands.
	
	\subsection{Privacy Protection for Industrial Data Sharing}
	Recent studies demonstrate that gradient information and smashed data can be exploited for reconstruction attacks. For instance, malicious attackers could infer product design specifications from vibration data features or reverse-engineer production line configurations through motion trajectory analysis.
	Future research directions regarding industrial data security encompass: 1) Developing hybrid encryption frameworks combining homomorphic encryption for gradient protection and differential privacy for smashed data obfuscation, with adaptive noise injection based on data sensitivity levels. 2) Creating attack-resistant FSL frameworks through feature space randomization and dimension compression.
	3)~Another promising method involves integrating transformer-based attention mechanisms to dynamically mask sensitive feature channels during transmission.
	
	\subsection{Human-in-the-Loop Collaboration for Industrial Robots}
	Although FSL improves robotic flexibility and autonomy, effective human-robot collaboration remains essential in complex manufacturing scenarios. Current systems lack: i) user-friendly interfaces for workers to monitor and correct robot operations, ii) methods to integrate expert knowledge into model updates, and iii) safety measures to ensure secure human involvement.
	Research directions regarding human-in-the-loop collaboration include: 1) Developing augmented reality (AR) interfaces for visualizing industrial manufacturing, enabling workers to identify and correct biased operations. 2) Creating active learning frameworks where human feedback prioritizes uncertain data samples. 3) Designing safety-critical control systems that maintain human oversight in shared workspaces.
	
	
	\subsection{Energy-Aware FedSL for Sustainable Manufacturing}
	Smart factories are under increasing pressure to meet carbon-neutral goals. 
	FedSL, while communication-efficient, still induces non-negligible energy consumption due to (i) frequent radio transmissions and (ii) idle CPU/GPU waiting for stragglers.
	To transform FedSL into a sustainable learning framework suitable for next-generation smart factories, future research should include carbon-intensity-aware scheduling. This involves aligning training rounds with real-time grid carbon data so that heavy cloud updates are shifted to low-carbon hours. For instance, during high-carbon intensity periods, computations could be shifted toward on-site renewable-powered edge nodes to minimize the overall carbon footprint. Additionally, new optimization algorithms should be developed to balance model accuracy, training latency, and environmental impact. For example, reinforcement learning agents could be deployed at manufacturing cells to learn optimal scheduling strategies that minimize carbon emissions while meeting production deadlines.
	
	\section{Conclusions}
	This article presents a comparative analysis of various FedSL frameworks applied within industrial IoT environments involving robots with limited hardware and communication resources.
	Specifically, four representative FedSL frameworks, such as synchronous, asynchronous, hierarchical, and heterogeneous FedSL, are investigated from the aspects of their key characteristics, operational workflows, as well as their respective advantages and limitations.
	Furthermore, three token fusion strategies are categorized: pre-fusion, intra-fusion, and post-fusion.
	The discussion then extends to adaptive optimization techniques employed within FedSL frameworks, including model compression, split layer selection, computing and communication resource allocation.
	Simulation results evaluate the efficacy of FedSL frameworks in industrial scenarios with resource-limited robots.
	Finally, open issues and future research directions are presented at the nexus of FedSL and industrial robotics.


	\bibliographystyle{IEEEtran}
	\bibliography{IEEEabrv, ref}

\end{document}